\documentclass[11pt]{article}
\usepackage[margin=1in]{geometry}
\usepackage{url,hyperref,microtype,subcaption,graphicx,wrapfig,amsmath,amssymb}
\usepackage[onehalfspacing]{setspace}
\usepackage{acronym}
\usepackage{natbib} 
\acrodef{DL}[\textsc{DL}]{deep learning}
\acrodef{rov}[\textsc{ROV}]{remotely operated vehicle}

\title{Is AI currently capable of identifying wild oysters? A comparison of human annotators against the AI model, ODYSSEE}

\author{
Brendan Campbell$^{1,*}$, 
Alan Williams$^{2,*}$, 
Kleio Baxevani$^{1,3,*}$, 
Alyssa Campbell$^{1,4}$, \\
Rushabh Dhoke$^{5}$, 
Rileigh E. Hudock$^{1}$, 
Xiaomin Lin$^{6,7}$, 
Vivek Mange$^{5,6}$, \\
Bernhard Neuberger$^{8}$, 
Arjun Suresh$^{6}$, 
Alhim Vera$^{9}$, 
Arthur Trembanis$^{1}$, \\
Herbert G. Tanner$^{5}$, 
Edward Hale$^{1,4}$
}

\date{}

\begin{document}
\maketitle

\begin{center}
$^{1}$University of Delaware, College of Earth, Ocean, and Environment, School of Marine Science and Policy, Lewes, DE, USA \\
$^{2}$University of Maryland Center for Environmental Science, Horn Point Laboratory, Cambridge, MD, USA \\
$^{3}$University of Maryland, Department of Aerospace Engineering, College Park, MD, USA \\
$^{4}$University of Delaware, College of Earth, Ocean, and Environment, Delaware Sea Grant, Lewes, DE, USA \\
$^{5}$University of Delaware, Center for Autonomous and Robotic Systems, Newark, DE, USA \\
$^{6}$University of Maryland, Maryland Robotics Center, College Park, MD, USA \\
$^{7}$Johns Hopkins University, Institute for Assured Autonomy, Baltimore, MD, USA \\
$^{8}$University of Applied Sciences Technikum Wien, Department of Industrial Engineering, Vienna, Austria \\
$^{9}$University of Cincinnati, Department of Aerospace Engineering and Engineering Mechanics, Cincinnati, OH, USA \\
\vspace{1em}
$^*$Corresponding authors: \texttt{bpc@udel.edu}, \texttt{awilliams@umces.edu}, \texttt{kleio@umd.edu}
\end{center}

\vspace{2em}
\begin{abstract}
Oysters are ecologically and commercially important species that require frequent monitoring to track population demographics (e.g. abundance, growth, mortality). Current methods of monitoring oyster reefs often require destructive sampling methods and extensive manual effort. Therefore, they are suboptimal for small-scale or sensitive environments. A recent alternative, the ODYSSEE model, was developed to use deep learning techniques to identify live oysters using video or images taken in the field of oyster reefs to assess abundance. The validity of this model in identifying live oysters on a reef was compared to expert and non-expert annotators. In addition, we identified potential sources of prediction error. Although the model can make inferences significantly faster than expert and non-expert annotators (39.6 s, $2.34 \pm 0.61$ h, $4.50 \pm 1.46$ h, respectively), the model overpredicted the number of live oysters, achieving lower accuracy (63\%) in identifying live oysters compared to experts (74\%) and non-experts (75\%) alike. Image quality was an important factor in determining the accuracy of the model and the annotators. Better quality images improved human accuracy and worsened model accuracy. Although ODYSSEE was not sufficiently accurate, we anticipate that future training on higher-quality images, utilizing additional live imagery, and incorporating additional annotation training classes will greatly improve the model's predictive power based on the results of this analysis. Future research should address methods that improve the detection of living vs. dead oysters.
\end{abstract}

\vspace{1em}

\section{Introduction} \label{sec:intro}

Globally, historical oyster reefs decreased by approximately 85\% \citep{beck2011oyster}. Reductions in global oyster populations were attributed to several factors, including disease \citep{Haskin1982,Andrews1988,Petton2021}, overharvest \citep{Rothschild1994} and climate change \citep{McFarland2022,Heo2023,Neokye2024}. Since oysters are ecosystem engineers, the reduced reef area and continuity affect the beneficial services provided to other aquatic organisms, resulting in a disproportional impact on marine communities \citep{Rodney2006,Grabowski2022}. 

The formation of oyster reefs increases the rugosity of the seafloor, producing shear stress that attenuates wave action and downstream velocity \citep{Kitsikoudis2020,Campbell2025} and stabilizes sediments \citep{Scyphers2011,Peyre,Walles2015}. The structures also provide valuable habitat for many fishes and invertebrates \citep{Harding2001,Connolly2024}. Furthermore, oyster filtration was historically capable of filtering entire estuarine regions \citep{Wiltsee, ZuErmgassen2013}, removing suspended particles, facilitating benthic coupling \citep{hoellein,Testa2015,Ray2021}, and improving water clarity \citep{Newell2004}. Oysters are also culturally and commercially significant for many coastal communities \citep{michaelis}. In the United States alone, wild oyster harvest was valued at \$221 million in 2019 \citep{noaa}, and global oyster aquaculture grows roughly 2\% annually \citep{Botta2020}. 

The importance of oyster reefs for ecosystem and economic function has led to a global effort to restore natal beds \citep{Carranza2020}. These efforts often succeed in increasing the population of formerly eradicated reefs, revitalizing threatened species \citep{McAfee2024}, and expanding the reef area \citep{Hernández}, without affecting genetic diversity \citep{Hornick2022}. Furthermore, restored reefs provide ecosystem services similar to functionally undisturbed reefs \citep{Grabowski2022,Smith2024}, further elucidating the benefits of reef restoration efforts. However, while services are observed to improve during the first two years, the rate of service provision decreases over the following years, highlighting the need for long-term monitoring \citep{Hemraj2022}.  

Reef monitoring to manage local fisheries and assess habitat quality is often conducted annually to biannually, combining efforts from state and academic institutions with varying scales and objectives (e.g., New Jersey, \cite{NJdbay}; Delaware, \cite{DEDNREC}; Maryland, \cite{MDDNR}; Virginia, \cite{vosara}). Although monitoring is regionally specific and is directed to address local interests, many sampling methods and analytical metrics overlap. In Delaware, beds are dredged to collect roughly 1 bushel of reef material (35.2 L) to identify, count, and measure market-sized and small oysters. These individuals are used to derive size frequency distributions to determine the density of marketable oysters and understand reef population dynamics. Although these metrics provide valuable information on reef health and fisheries status, the processing time to conduct these studies in Delaware alone takes several days of boat time and the concentrated effort of six to seven trained crew members to sample a subset of the existing reefs (Audrey Ostroski, DNREC, personal communications). Furthermore, these assessments require destructive sampling methods, which cannot occur in small or vulnerable reef beds, hardened shoreline restoration projects, or newly seeded beds.

Modern advances in robotics and photogrammetry can facilitate data collection for oyster stock assessments while providing a minimally destructive alternative to traditional sampling methods. Stationary video cameras are often used to identify the abundance, composition, and behavior of nekton associated with coral reefs \citep{Boom2014}, oyster reefs \citep{Connolly2024}, and oyster aquaculture equipment \citep{Mercaldo,Armbruster2024,ambrose2024habitat}. Autonomous vehicles assessed the abundance and distribution of sea scallops in the Mid-Atlantic Bight \citep{Kannappan2014,Walker2016,rasmussen}. Aerial drones were successful in imaging the area and morphology of intertidal oyster reefs in North Carolina and Florida \citep{Windle2019,bennett2024assessing}. Hand-towed video cameras provided footage to assess localized habitat quality of harvested and non-harvested oyster beds \citep{Anchondo2024,Heggie2021}. The culmination of video-based surveying efforts in marine environments provides valuable information regarding habitat extent, quality, and composition, but at the species or individual level to address abundance or density, hours to days worth of data are required over longer time spans to acquire a sufficient dataset (e.g. \cite{Mercaldo, jensen2024migratory}). Furthermore, manual efforts to accurately assess video data require significant training and labor to process and quality control datasets, which is time-consuming, labor-intensive, and limits the scalability of research efforts \citep{English2024}. Fortunately, a catalog of more than 11 million benthic images is available open-access through BenthicNet to facilitate training and potential \ac{DL} applications, expanding opportunities for underwater image-based analyses \citep{lowe2025benthicnet}. 

Recent advancements on \ac{DL} have contributed to optimizing surveying methods in marine systems. \ac{DL}-based visual recognition and detection of aquatic animals combines advanced technologies such as image processing, computer vision, cloud computing, sonar, and sensors to provide automatic and intelligent identification, detection, and tracking of aquatic species \citep{fernandes2024artificial}. In addition, these techniques can be integrated into edge devices and directly help identify marine species, providing crucial support to researchers in the field \citep{lin2024odyssee}. Underwater robots can implement vision models in situ; however, training vision models often requires large datasets and manual annotations, which can be extremely difficult and tedious in coastal marine environments, limiting scalability \citep{huang2025artificial}. In recent years, advanced image generation techniques, such as stable diffusion, have allowed the generation of convincing synthetic data to enhance and diversify training data sets for detection models. This approach was used to train the OSYSSEE model to detect live oysters in a reef setting \citep{lin2024odyssee}.

In this study, we compared the performance of the DL oyster identification model, ODYSSEE, against expert and non-expert annotators to delineate potential sources of error associated with the identification of live oysters using photogrammetry methods. We anticipate that the model will accurately identify live oysters and potentially have limitations in its current stage in separating live and dead oysters. The main contribution of this work is to realize the potential sources of error in the model compared to human identifiers that can provide the necessary information to determine the validity of such models, a scientific monitoring tool, and inform pathways that can be used to enhance model performance. Then, utilizing \ac{DL} as a non-destructive means to census live oysters on a reef can significantly improve current monitoring efforts for stock assessments, restoration efforts, and aquaculture citing and harvest.





\section{Methods} \label{sec:method}

\subsection{Image Acquisition}
A set of 150 unique images was obtained using video footage captured on an oyster reef located in Lewes, DE USA (38.7890 N, 75.1624 W) during low tide, spanning multiple efforts between June and September 2024 when the water clarity was favorable. Data were recorded using a handheld camera system (drop camera method) with two GoPro Hero12 cameras mounted on a 2-meter-long PVC post or a \ac{rov} BlueROV2 heavy configuration (Blue Robotics, St. Torrance, CA, USA) with a GoPro Hero12 (GoPro, San Mateo, CA, USA) mounted on the payload skid. All images were unmodified with varied target distance, clarity, and number of oysters to represent the potential range of images available in a field setting. The quality score (QS) of each image was assessed based on previously published works \citep{Padole2019} to use a score of 0 - 4, where 0 indicates no visibility of desired features, 1 contains features that are present but are not interpretable, 2 contains features with limited or inconsistent quality, 3 indicates acceptable quality for most of the image and 4 indicates exceptional quality throughout the entire image. Images with a QS of 0 or 1 were not used in the study. 

\subsection{Deep Learning Model}
The best-performing model from \cite{lin2024odyssee} (highest recorded mAP 0.657, called ODYSSEE) was developed using the YOLOv10 model platform and trained on a dataset of 30\% human-annotated data of oysters in the wild and 70\% synthetically generated images to provide an efficient source of data using methods by \cite{lin2022oystersim} to render oysters based on 3D scans of live specimens. These renderings were then passed through a stable diffusion model \citep{rombach2022highresolutionimagesynthesislatent} in conjunction with various ControlNets \citep{zhang2023addingconditionalcontroltexttoimage} to improve the generated output and more effectively bridge the sim2real gap.

\subsection{Annotation Process}

To compare the accuracy and detection rate across potential end users, 150 images were analyzed by (i) the oyster detection model, (ii) 5 expert annotators (biologists who frequently work with oysters), and (iii) 5 non-expert annotators (deemed by having minimal hands-on experience with oysters outside of this project). All human annotators are listed as coauthors, and no annotations were completed by a third party. The model identifies live oysters with a confidence score (CS) and only accounts for CS greater than 0.5 (potential range = 0 - 1). Non-expert and expert annotators identified live (confident identification of a live individual), dead (confident identification of a dead individual or loose shell), or unknown (confident identification of an oyster but not confident whether live or dead). Manual annotations were made using RoboFlow \citep{roboflow} using preset options for each observation class. Annotators drew a bounding box around their observations and marked them with the appropriate class. The program then recorded the number of identifications per class, per image. The images were annotated in the same order and the annotators were not encouraged to review the completed images to standardize the exposure time between all the annotators for each image. The final time to annotate all 150 images was also recorded. 

\subsection{Analysis of Model Performance}
Across each annotator treatment, pairwise tests were used to determine the average number of live identifications made per image and assess the differences in time needed to complete the annotations, considering an alpha of 0.05. To determine consistency, the intraclass correlation coefficient (ICC) was calculated between and within the annotator groups per image. When comparing annotator treatments with unequal sample sizes, the identifications were averaged and rounded to standardize the range of potential outcomes between the groups.

A random subset of 30 images was evaluated for all unique identifications against all annotators to determine the agreement between the individual oysters observed. The 'true' identification was determined in two ways: the first was to account for all potential outcomes, and the second was to consider only live identifications. For the first method, when two or fewer annotators identified an object, we denoted that observation as a false positive. When three or more groups identified the object, we determined that the true value was the observation made the most frequently. If the number of live classifications was equal to dead or unknown, the object would be considered live. If the number of dead classifications was equal to unknown, the object would be considered dead. For the second method, any observation without live identifications was removed from the dataset. Any object with two or fewer live identifications was classified as a false positive and anything with three or more live identifications was considered live. For both methods, a CS was generated for each result by dividing the total number of classifications that matched the true result by the total number of times that object was identified. The confidence matrices were then created to understand the agreement between the individual identifications and to describe the sources of error between the groups of annotators. For live or no-observation evaluation, the resulting outcomes from the confusion matrices were used to derive Area Under Curve (AUC) values to determine false detection rates from each group when identifying live oysters. 

\section{Results}
\begin{figure}[h!]
       \centering
       \includegraphics[scale=0.6]{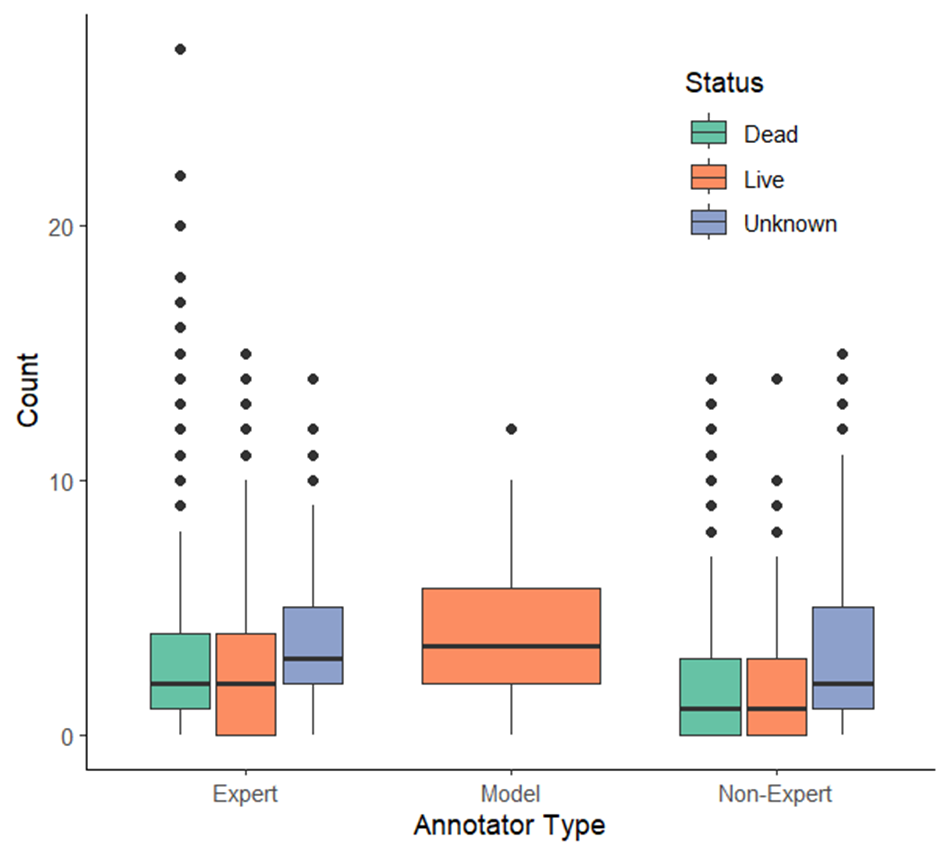}
    \caption{The model appears to over-predict live (orange) oysters compared to expert and non-expert annotators. Expert annotators also make a greater number of dead (green) and unknown (purple) annotations per figure compared to non-experts. The model treatment has a wider box for live annotations since it only has the capability of making live identifications.}
    \label{fig:FIGURE1.png}
   \end{figure}

    Of the 150 images used, 54 had a QS of 2 followed by 75 with a QS of 3 and 21 with a QS of 4, following a non-normal distribution ($p< 0.001$, Shapiro-Wilk test). The model annotated 150 images in 39.6 s, which is negligible compared to experts ($2.34 \pm 0.61$ h, n = 5; average ± standard deviation), which were significantly faster than non-experts ($4.50 \pm 1.46$ h, n = 5; $p = 0.021$, Wilcoxon signed-rank test). Across all images, the model made an average of 4 ± 2.45 live observations per image, which was greater than experts (2.76 ± 3.20), and non-experts (1.89 ± 2.08; $p < 0.001$, Kruskal-Wallis and Dunn test for all combinations). The model generated an average CS of 0.65 ± 0.18 across all its live observations. Experts made more dead (3.12 ± 3.57) and unknown (3.49 ± 2.62) observations than non-experts (2.04 ± 2.53, 4.69 ± 2.85, respectively; $p< 0.001$, Wilcoxon signed-rank test, Fig. \ref{fig:FIGURE1.png}).

\begin{figure}[h!]
       \centering
       \includegraphics[scale=0.375]{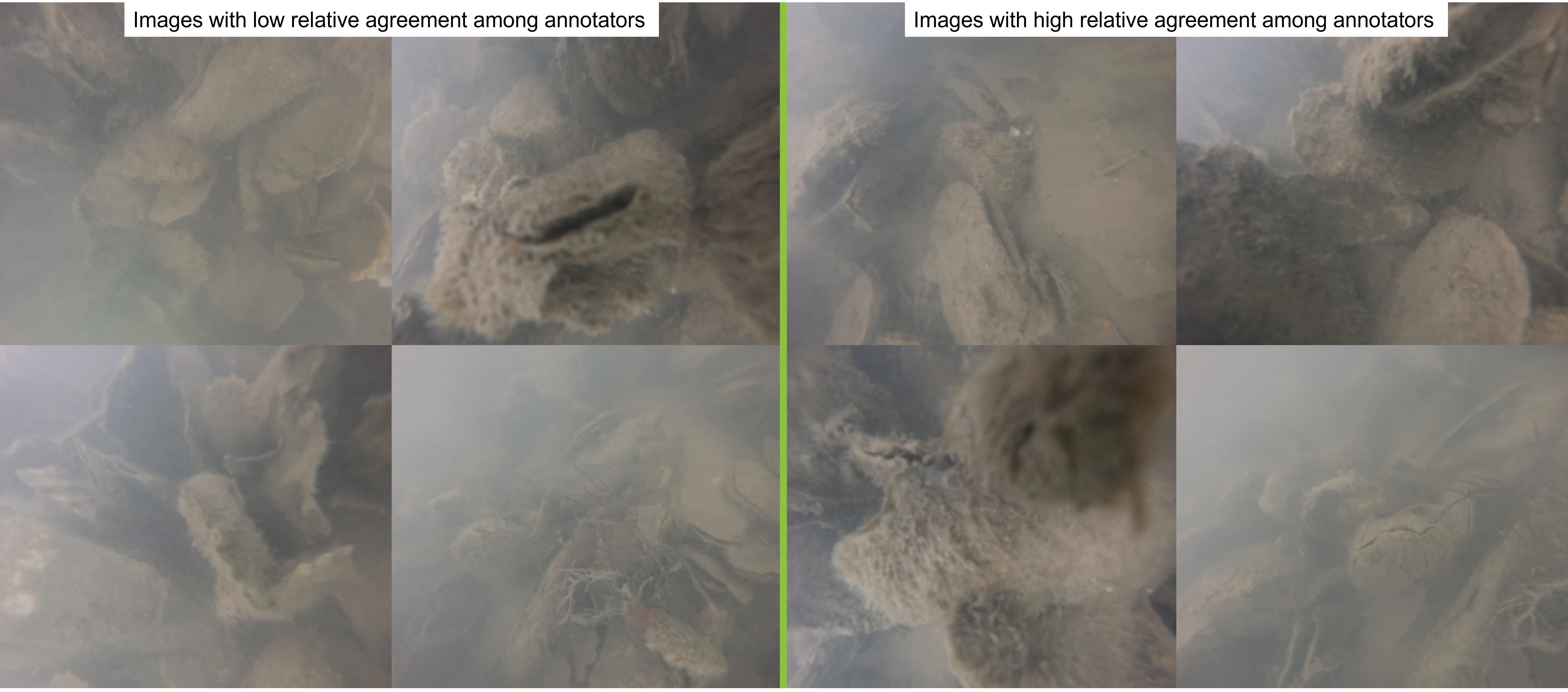}
    \caption{Examples of images used that represented low (left) and high (right) relative agreement between all annotators.}
    \label{fig:AgreeabilityImage.png}
   \end{figure}

	At the image scale, the consistency of live detections between all annotator groups for an image was poor (ICC = 0.430, $p < 0.001$) and was moderate between experts and non-experts (ICC = 0.673, $p < 0.001$, Fig.\ref{fig:AgreeabilityImage.png}). Similarly, the consistency of the experts and non-experts in observing dead and unknown detections in each image was also poor (ICC = 0.379, $p < 0.001$; ICC = 0.282, $p < 0.001$; respectively). However, the consistency of observations increased with better image quality. The consistency of live observations across all annotator groups increased from 0.366 at QS = 2 to 0.420 at QS = 4. Expert and non-expert groups had poor consistency of live observations at QS = 2 and moderate consistency at QS = 3 and QS = 4 (ICC = 0.466, 0.696, 0.658, respectively, $p < 0.001$). For dead observations, images with a QS of 2 and 3 had moderate consistency and good consistency at QS = 4 (ICC = 0.528, 0.657, 0.890, respectively, $p < 0.001$). However, the consistency of unknown observations decreased with QS between expert and non-expert annotators (ICC when QS = 2, 0.649; when QS = 3, 0.540, $p < 0.001$) showing poor consistency at QS = 4 (ICC = 0.226, p = 0.156). 
    
\begin{figure}[b!]
    \centering
    \includegraphics[width=0.3\textwidth]{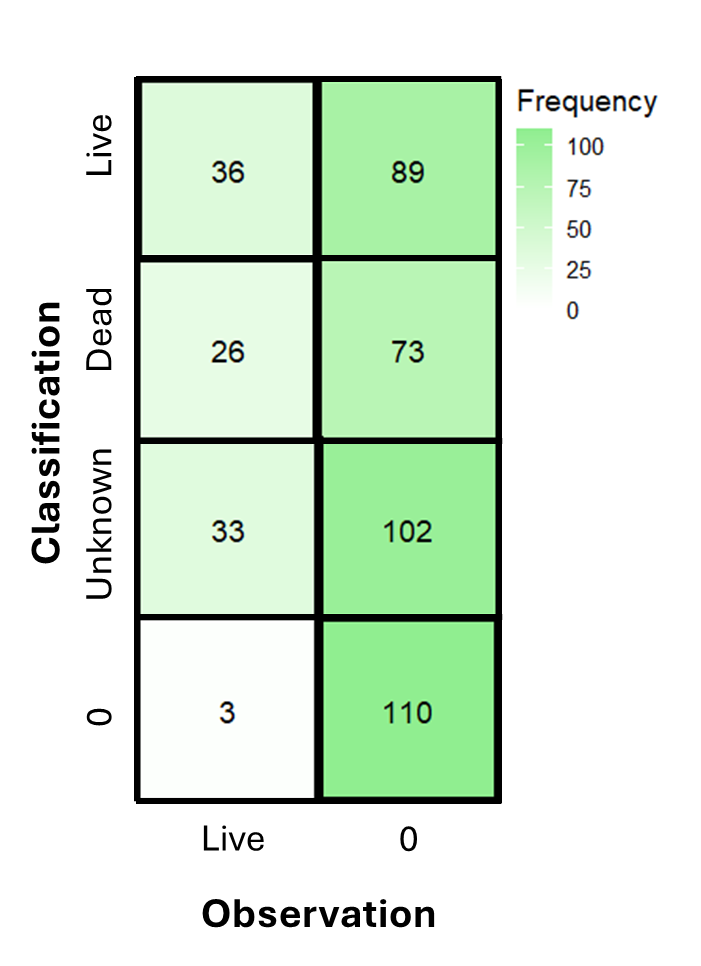}
            \caption{Confusion matrix from the model demonstrates that the majority of classifications were not identified by the model, primarily `unknown' oysters. A classification or observation of `0' denotes a false positive or missed observation.}
        \label{fig:conf_matrices_model}
    \end{figure}
\begin{figure}[h!]
        \centering
        \includegraphics[scale=0.46]{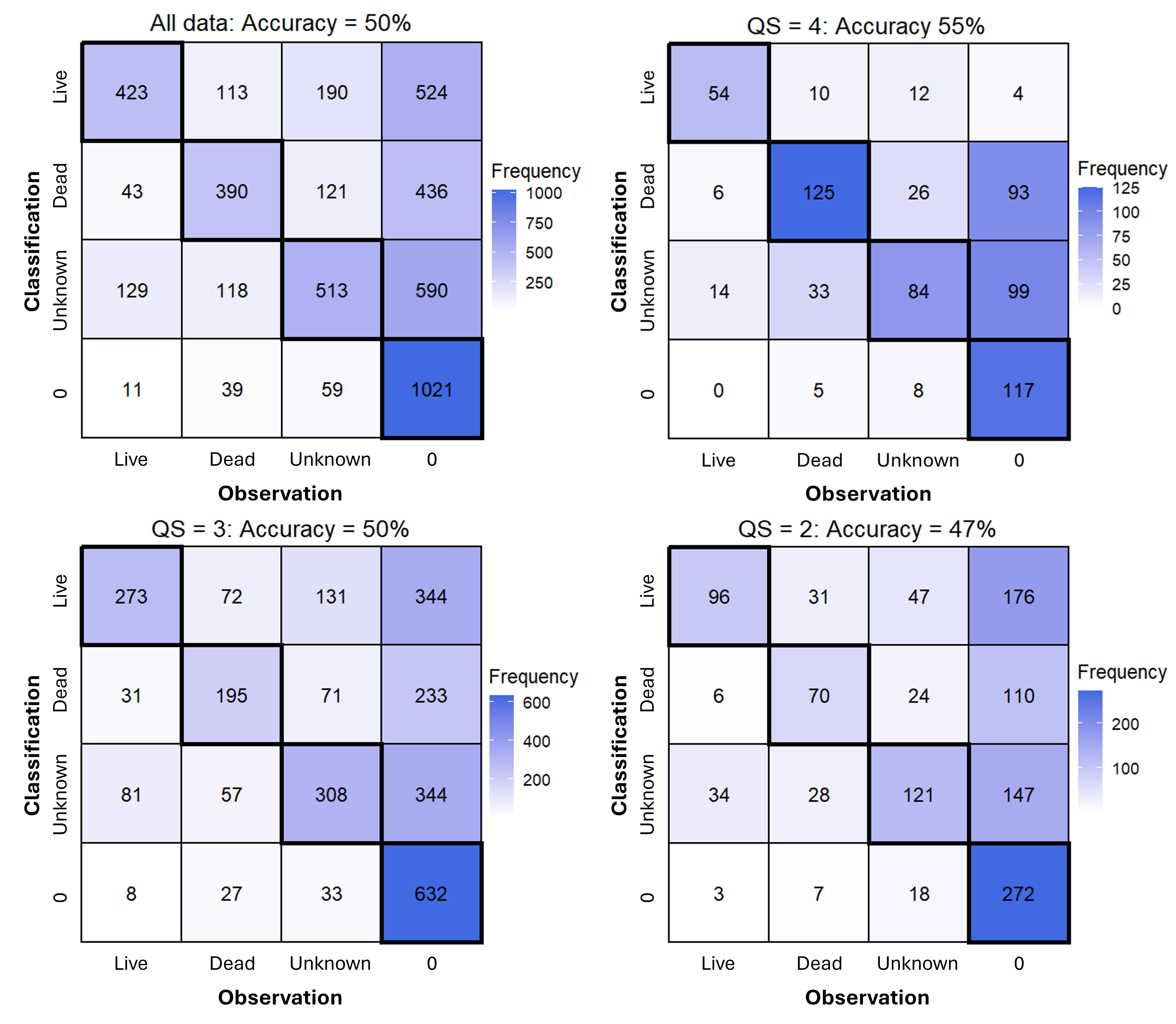}
        \caption{Confusion matrices from all data across expert and non-expert annotators (top left) and separated by QS show an increase in prediction accuracy and reduction in false positive detections (denoted as `0') with increasing QS.}
        \label{fig:conf_matrices}
    \end{figure}

	To determine the accuracy (Acc), Precision (Pre), and Recall (Rec) of each annotator group, 30 images were randomly selected to compare all unique observations made. From this subset, 125 live oysters (CS = 0.50 ± 0.17), 99 dead (CS = 0.57 ± 0.16), 135 unknown, and 113 false oyster detections were classified. The model made 36 correct observations, missed 89 live oysters, and misidentified 26 dead, 33 unknown, and 3 false-positive oysters as live (Fig. \ref{fig:conf_matrices_model}). Experts, who classified an average of 72.2 ± 67.5 live oysters, positively identified 39\% of the live oysters. Non-experts, who classified 49 ± 24.5 live oysters, only positively identified 27\% of the live oysters. Of the average 76 ± 63.8 dead observations made by experts and 56 ± 47.8 by non-experts, 46\% and 33\% of dead oysters were positively categorized, respectively. When comparing all independent responses and results from expert and non-expert annotators, accurate and false negative classifications were most often observed, regardless of image QS, however, accuracy did increase positively with QS (Fig. \ref{fig:conf_matrices}). Images with a QS = 2 resulted in 47\% Acc (F1 = 0.39, Pre = 0.27, Rec = 0.69), while images with a QS = 3 resulted in 50\% Acc (F1 = 0.45, Pre = 0.33, Rec = 0.69) and images with a QS = 4 resulted in 55\% Acc (F1 = 0.70, Pre = 68, Rec = 0.73).  
    
\begin{figure}[h!]
        \centering
        \includegraphics[scale=0.46]{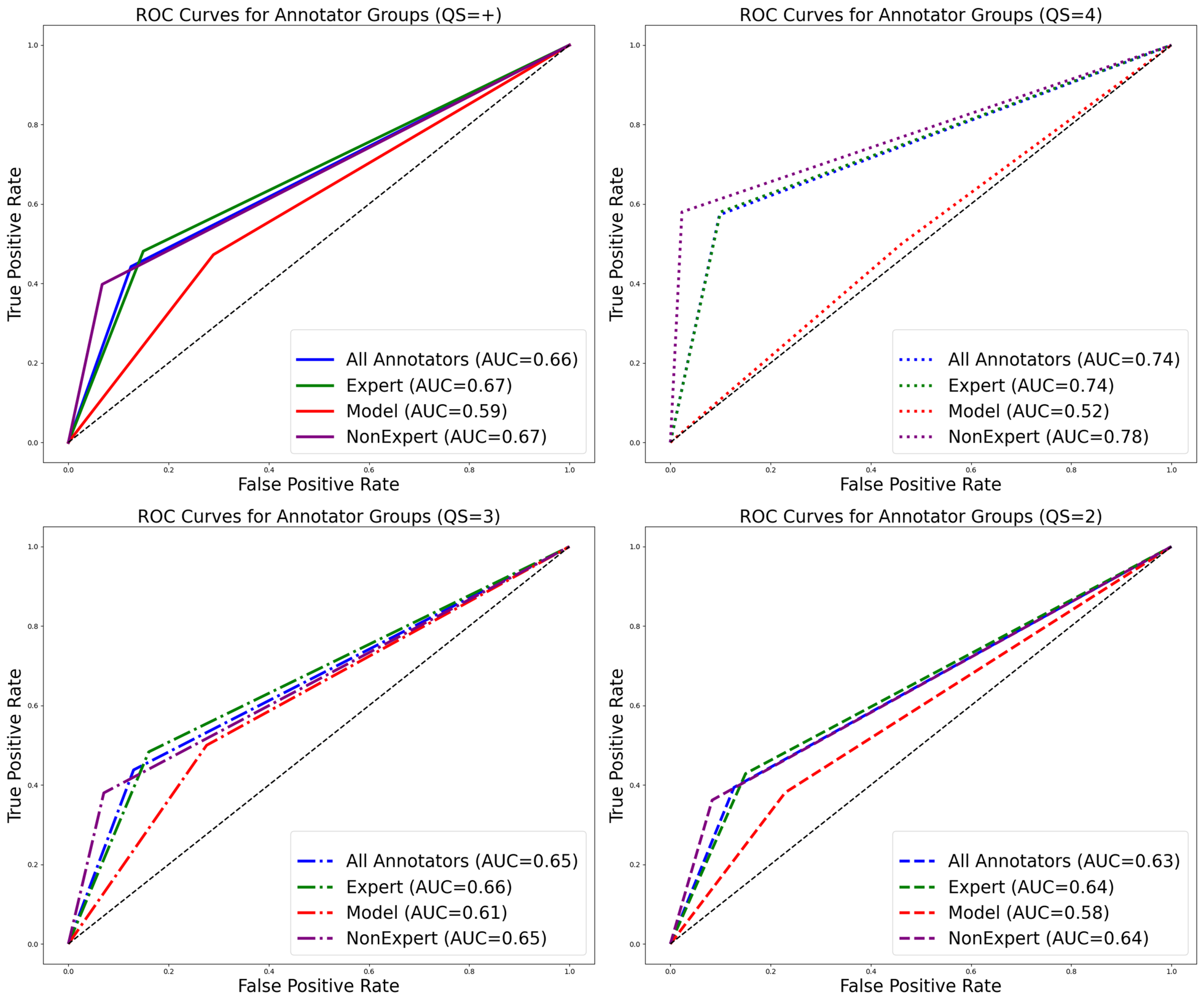}
        \caption{Receiver Operating Characteristic (ROC) curve plots including the dataset where all observations and classifications were simplified to `live' or `not observed' for all annotator groups. Plots and line types are separated by quality score and color is separated by annotator group. Data including all QSs is represented in the top left plot. Higher area under the curve values observed with increasing QS for expert and non-expert annotators and lower area under the curve values observed for the model with increasing QS.}
        \label{fig:ROC-split.png}
    \end{figure}

	To standardize outcomes between the model and human annotators, we simplified all observations and classifications to ‘live’ or ‘not observed.’ Using these criteria in the independent observations of the 30 images subsampled, 91 live oysters (CS = 0.41 ± 0.18) and 190 false detections were made. The model made 98 live observations and 47\% were correct. Experts with an average 72.2 ± 67.5 live observations accurately predicted 48\% of live classifications and non-expert made 49 ± 24.5 live observations and were correct 40\% of the time. From the confusion matrices, when all annotator groups were pooled, there was 74\% accuracy among correct or true negative observations (live detections, F1 = 0.52, Pre = 0.44, Rec = 0.63). Experts (Acc = 73\%; live detections, F1 = 0.54, Pre = 0.48, Rec = 0.61) and non-experts (Acc = 76\%; live detections, F1 = 0.52, Pre = 0.40, Rec = 0.74) generally had improved performance compared to the model (Acc = 63\%; live detections, F1 = 0.46, Pre = 0.47, Rec = 0.44). The accuracy and receiver operating characteristics (ROC) are more ideal across higher QS for expert and non-experts and have the opposite effect on the model (Fig. \ref{fig:ROC-split.png}). Similarly, all annotator groups demonstrate a skew towards more false positive detections, with the model generating the largest deviance, followed by experts, the average across all annotators, then non-experts. 
    
\section{Discussion} \label{sec:discussion}


In this study, we evaluated the ability of a developed DL model to distinguish and quantify living oysters from their surroundings accurately. We then compared the performance of the DL model to observations made by expert and non-expert annotators. Accurately identifying oysters using automated detection models would greatly improve the efficiency and cost of reef surveying, permitting greater scalability. While the image processing time was immediate and can run in situ \citep{lin2024odyssee}, the model greatly over-predicted the presence of live oysters with poor consistency (ICC = 0.430, $p < 0.001$), lower accuracy, and higher false positive rates (Acc = 63\%, AUC = 0.59) compared to expert (Acc = 73\%, AUC = 0.67) and non-expert annotators (Acc = 76\%, AUC = 0.67). Between human groups, experts annotated images almost twice as fast. They made more accurate observations than non-experts, but had a greater false negative rate and a less balanced skew between accurate and false detections. There was a considerable difference in accuracy with image quality, highlighting a gap in the current training process for the model. Human annotators had an overall increasing accuracy in detecting live and dead oysters as image quality increased, while the model was less accurate when image quality increased.

The ODYSSEE model provides a valuable estimate of living oysters in reefs by identifying the presence of live and dead animals within a dense aggregation of shells, a functional need for an abundant reef to be present. However, a stronger differentiation between live and dead oysters during the annotation process is required to properly quantify living oyster abundance and act as a reliable quantitative tool for oyster reef censusing for aquaculture, fisheries, and restoration applications. Still, estimating oyster shell density can be used for analysis in areas of large, living oyster aggregations where relative measures of abundance are sufficient. Having a relative abundance metric can be useful for on-bottom oyster farmers and the wild capture fishery to more efficiently site harvestable beds using imagery coupled with GPS data. The concept of constructing an oyster distribution map through ROV operation was simulated using the OysterSim model \citep{lin2022oystersim}, and when applying the proposed framework ShellCollect, by \cite{wang2024shellcollect}, the map can be used to generate an optimal harvesting path. The current stage of ODYSSEE likely provides sufficient context to develop harvest paths using these existing frameworks, and the continued improvements to ODYSSEE will better resolve the efficiency of the harvest paths generated. These maps can also be implemented in restoration efforts to determine locations to strategically replenish shell to support reef development and support other local fisheries \cite{marquardt2025oyster} by targeting areas of low relative oyster density. Beyond harvest path planning, alterations to the ODYSSEE model for pure detection of oyster shell substrate could prove valuable to site selection for both habitat restoration \citep{george2015oyster,hughes2023site} and oyster larval planting efforts \citep{spires2023direct}.  


Several considerations in our annotating procedures and data processing should be acknowledged for future studies. First, when annotating oysters in situ using photogrammetry methods, the presence of dead oysters that are forced shut from trapped sediment (referred to as `mudders') would likely be counted as a live oyster in any computer-based survey method, inflating the number of live oysters counted in a study compared to hand-grading methods, where manual assessments post-harvest can detect these methods by feeling for atypical density and differences in sound emitted when hit against a known live oyster. Another source of error could be attributed to the variable gaping behavior of oysters when they feed, which may be indistinguishable from dead oysters, providing a source of disagreement between annotators and potential error when counting live individuals. Although ICC analyses provide some context and can establish relative consistency through future model iterations, manually analyzing all unique observations across all annotators provides a more precise assessment of consistency since ICC metrics rely on perfect agreement from an image scale rather than an individual identification scale, although manually checking unique observations for a more in-depth analysis is time-consuming and tedious. 

To improve model performance, the number of false negative and false positive observations need to be reduced and consistency between the model and human annotators needs to increase. Qualitatively, most false positives were caused by rocks, mussels, and shadows. A potential avenue is to use a larger percentage of live imagery in our training datasets. ODYSSEE is currently trained using 30\% real oyster images and 70\% synthetic oyster images \citep{lin2024odyssee}. Although incorporating synthetic data helps to provide a larger training dataset, it can overlook features of oysters noticeable in natural imagery, reducing precision \citep{nowruzi2019much,corvi2023detection,wang2017effectiveness}. Furthermore, providing additional higher-quality imagery and determining an intentional distribution of image quality to the existing training dataset would likely improve model performance by providing more variable content for the model to learn from. Furthermore, several annotators should annotate real images, and the associated CS classification should provide context for clear and less clear observations \citep{sullivan2018deep}. Lastly, providing additional annotations to the detection model (i.e. dead, unknown, and other objects such as mussels, rocks, and debris), and better quantifying sources of false positive detections can be a possible solution to minimize false positive detections \citep{shorten2019survey}. 

While comparing model accuracy to expert and non-expert annotators provides some context on model performance using a relative measure of a live oyster, a more valid ground-truth with known live oysters would be a more insightful test of model performance. A comparison to hand-counted living oysters in a reef setting would be an ideal method for making model accuracy assessments. However, such an assessment would be difficult and costly to perform. This methodology is frequently used in terrestrial landscapes, although there are considerably fewer logistical barriers \citep{shah2024leveraging}. 

Our study demonstrates the potential of DL models to automate oyster detection from underwater imagery, offering a non-invasive alternative to traditional, labor-intensive monitoring methods like dredging. By integrating real and synthetic data for training, we reduce reliance on destructive sampling while improving the efficiency and accessibility of reef assessments under variable conditions. These advances have broad applications, benefiting restoration and fisheries management efforts, where frequent and minimally invasive monitoring is essential. This approach has promise for oyster farmers looking to utilize technology to reduce labor demands and streamline aquaculture operations. Using tools that reduce processing time provides a foundation for sustainable reef management while supporting the growing intersection of aquaculture and technology. 


\include{Sections/Conclusion}

\section*{Author Contributions}

BC: Conceptualization, Data curation, Formal Analysis, Investigation, Methodology, Project Administration, Validation, Visualization, Writing - original draft; 
AW: Conceptualization, Formal Analysis, Investigation, Methodology, Project administration, Resources, Writing - original draft; 
KB: Conceptualization, Data curation, Investigation, Methodology, Project administration, Resources, Software, Supervision, Writing - review \& editing; 
AC: Investigation, Writing - review \& editing; 
RD: Formal Analysis, Investigation, Visualization, Writing - review \& editing;
RH: Data curation, Investigation, Resources, Writing - original draft; 
XL: Conceptualization, Data curation, Methodology, Project administration, Resources, Software, Writing - review \& editing; 
VM: Conceptualization, Methodology, Writing - review \& editing; 
BN: Conceptualization, Investigation, Writing - review \& editing; 
AS: Data curation, Investigation, Methodology, Resources, Writing - original draft; 
AV: Conceptualization, Investigation, Writing - review \& editing; 
AT: Funding acquisition, Resources, Supervision, Writing - review \& editing; 
HT: Funding acquisition, Resources, Supervision, Writing - review \& editing; 
EH: Funding acquisition, Investigation, Resources, Supervision, Writing - review \& editing


\section*{Funding}
 This material is based upon work supported by the US Army Corps of Engineers, ERDC Contracting Office under Contract No. W912HZ-22-2-0015, "Transforming Shellfish Farming with Smart Technology and Management Practices for Sustainable Production" grant no. 2020-68012-31805/project accession no. 1023149 from the USDA National Institute of Food and Agriculture, and NOAA's Project ABLE via award number NA22OAR4690620-T1-01. Any opinions, findings, conclusions, or recommendations expressed in this material are those of the author(s) and do not necessarily reflect the views of the US Army Corps of Engineers, ERDC Contracting Office or the U.S. Department of Agriculture.

\section*{Acknowledgments}
We would like to acknowledge the various funding agencies that supported this work and the Autonomous Systems Bootcamp (ASB) for providing the opportunity for this global interdisciplinary collaboration. Furthermore, we thank Matthew Gray and Yiannis Aloimonos for their support and mentorship and Maxwell Collins for providing technical support during fieldwork and in processing some of the preliminary imagery.


\section*{Data Availability Statement}
The datasets generated for this study, including images and summary data can be found in the GitHub repository, 'OysterDetectionValidation' [https://github.com/Campbellb13-UD/OysterDetectionValidation/tree/main]. Larger files can be made available per request to the corresponding authors.

\bibliographystyle{Frontiers-Harvard} 

\bibliography{frontiers}


\end{document}